\documentclass[runningheads]{llncs}
\usepackage[T1]{fontenc}
\usepackage{graphicx}
\usepackage{amsmath,amssymb,amsfonts}
\usepackage{algorithmic}
\usepackage{graphicx}
\usepackage{textcomp}
\usepackage{xcolor}
\usepackage{hyperref}
\usepackage{booktabs}
\usepackage{adjustbox}
\usepackage{tabularx}
\usepackage{textcomp}
\usepackage{xcolor}
\usepackage{hyperref}
\usepackage{multirow}
\usepackage{graphicx}
\usepackage{amsmath}
\usepackage{booktabs}
\usepackage{tikz}
\usepackage{changepage}
\usepackage{physics}
\usepackage{mathdots}
\usepackage{yhmath}
\usepackage{amsfonts}
\usepackage{cancel}
\usepackage{color}
\usepackage{siunitx}
\usepackage{array}
\usepackage{multirow}
\usepackage{amssymb}
\usepackage{gensymb}
\usepackage{tabularx}
\usepackage{extarrows}
\usepackage{wrapfig}
\usepackage{xcolor}
\usetikzlibrary{fadings}
\usetikzlibrary{patterns}
\usetikzlibrary{shadows.blur}
\usetikzlibrary{shapes}
\usepackage{caption}
\usepackage{comment}

\def\BibTeX{{\rm B\kern-.05em{\sc i\kern-.025em b}\kern-.08em
    T\kern-.1667em\lower.7ex\hbox{E}\kern-.125emX}}

\usepackage{acronym}
\newacro{gan}[GAN]{generative adversarial network}
\newacro{vae}[VAE]{variational autoencoder}
\newacro{cnn}[CNN]{convolutional neural network}
\newacro{dire}[DIRE]{diffusion reconstruction error}
\newacro{mlp}[MLP]{multi-layer perceptron}
\newacro{clip}[CLIP]{contrastive language image pre-training}
\newacro{vlm}[VLM]{vision-language model}
\newacro{vit}[ViT]{vision transformer}
\newacro{nlp}[NLP]{natural language processing}
\newacro{cv}[CV]{computer vision}
\newacro{blip}[BLIP]{bootstrapping language image pre-training}
\newacro{vqa}[VQA]{visual question answering}
\newacro{lora}[LORA]{low-rank adaptation}
\newacro{peft}[PEFT]{parameter-efficient fine-tuning}
\newacro{gpt}[GPT]{generative pre-trained transformer}
\newacro{q-former}[Q-Former]{querying transformer}
\newacro{sedid}[SeDID]{stepwise error for diffusion-generated image detection}
\newacro{sd}[SD]{stable diffusion}
\newacro{lsun}[LSUN]{large-scale scene understanding}
\newacro{fc}[FC]{fully-connected}
\newacro{deit}[DeiT]{data-efficient image transformers}
\newacro{ldm}[LDM]{latent diffusion model}
\newacro{adm}[ADM]{ablated diffusion model}
\newacro{ddpm}[DDPM]{denoising diffusion probabilistic models}
\newacro{iddpm}[IDDPM]{improved denoising diffusion probabilistic models}
\newacro{pndm}[PNDM]{pseudo numerical methods for diffusion models on manifolds}
\newacro{srm}[SRM]{spatial rich model}
\newacro{lasted}[LASTED]{language-guided synthesis detection}
\newacro{rf}[RF]{random forest}
\newacro{dm}[DM]{diffusion model}
\newacro{ddim}[DDIM]{denoising diffusion implicit models}
\newacro{multilid}[multiLID]{multi local intrinsic dimensionality}
\newacro{ifdl}[IFDL]{image forgery detection and localization} 
\newacro{svm}[SVM]{support vector machine} 
\newacro{ai}[AI]{artificial intelligence} 

\newacro{amsff}[AMSFF]{attention-based multi-scale feature fusion} 
\newacro{psm}[PSM]{patch selection module}
\newacro{llm}[LLM]{large language model}
\renewcommand{\thetable}{\arabic{table}}

\begin{document}
\title{SAViL-Det: Semantic-Aware Vision-Language Model for
Multi-Script Text Detection\thanks{The supports of TotalEnergies and Sorbonne University Abu Dhabi are fully acknowledged.}}
\titlerunning{SAViL-Det}

\author{Mohammed-En-Nadhir Zighem\inst{1} \and
Abdenour Hadid\inst{1}}
\authorrunning{ Zighem, Hadid}
\institute{Sorbonne Center for Artificial Intelligence, Sorbonne University Abu Dhabi, UAE}
\maketitle              % typeset the header of the contribution
\begin{abstract}
Detecting text in natural scenes remains challenging, particularly for diverse scripts and arbitrarily shaped instances where visual cues alone are often insufficient. Existing methods do not fully leverage semantic context. This paper introduces SAViL-Det, a novel semantic-aware vision-language model that enhances multi-script text detection by effectively integrating textual prompts with visual features. SAViL-Det utilizes a pre-trained CLIP model combined with an Asymptotic Feature Pyramid Network (AFPN) for multi-scale visual feature fusion. The core of the proposed framework is a novel language-vision decoder that adaptively propagates fine-grained semantic information from text prompts to visual features via cross-modal attention. Furthermore, a text-to-pixel contrastive learning mechanism explicitly aligns textual and corresponding visual pixel features. Extensive experiments on challenging benchmarks demonstrate the effectiveness of the proposed approach, achieving state-of-the-art performance with F-scores of 84.8\% on the benchmark multi-lingual MLT-2019 dataset and 90.2\% on the curved-text CTW1500 dataset. %The code will be publicly available upon acceptance. 

\end{abstract}

%
%
%
%%%%%%%%%%%%%%%%%%%%%%%%%%%%%%%%%%%%%%%%%%%%%%%%%%%%%%%%%%%%%%%%%%%%
\vspace{-6mm}
\section{\textbf{Introduction}}
\label{sec:intro}
%\vspace{-2mm}

Scene text detection, the task of localizing text within natural images, is a fundamental challenge in computer vision with wide-ranging applications. While significant progress has been made, accurately detecting text remains difficult, especially in complex "in-the-wild" scenarios. These challenges are amplified when dealing with diverse scripts (e.g. multi-lingual text) and irregular text shapes, such as curved or arbitrarily oriented text, where visual cues alone can be ambiguous or insufficient. 

Existing approaches to open-scene text detection \cite{raisi2020text,blanco2022survey} often fall into regression based or segmentation based categories. Regression methods predict text geometry directly, evolving to handle complex shapes, while segmentation methods treat it as a pixel classification problem. Some methods combine detection and recognition, use scale expansion, differentiable binarization, or pixel linking. More recently, cross-modal approaches have emerged, integrating information beyond visual input, like textual content within the image, often leveraging powerful pre-trained models like CLIP (Contrastive Language-Image Pre-training \cite{radford2021learning}). These methods aim for more robust detection by combining diverse information sources. However, most existing cross-modal techniques do not fully exploit the potential of textual prompts or rely heavily on self-attention mechanisms within visual processing, potentially overlooking valuable semantic guidance. To address these limitations, this paper proposes SAViL-Det, a novel semantic-aware vision-language model for multi-script text detection. Our approach leverages the power of pre-trained vision-language models (CLIP)  and introduces a new cross-modal pre-training paradigm. The core 
 of our proposed framework is a language-vision decoder designed to adaptively propagate fine-grained semantic information from textual prompts to visual features. This integration is further enhanced by an Asymptotic Feature Pyramid Network (AFPN)~\cite{yang2023afpn} for robust multi-scale feature fusion  and a text-to-pixel contrastive learning mechanism to explicitly align textual and visual representations at the pixel level. 
 
 Our main contributions are:
(1) A novel language-vision decoder that effectively integrates semantic information from text prompts into the visual feature extraction process. (2) A cross-modal framework (SAViL-Det) employing CLIP, AFPN, and text-to-pixel contrastive learning for enhanced text localization. (3) State-of-the-art performance on challenging multi-lingual (MLT-2019)~\cite{MLT2019} and curved-text (CTW1500)~\cite{CTW1500} benchmarks.
This paper describes our proposed methodology, including the feature extraction, fusion, and cross-modal decoding processes. We present comprehensive experimental results, including an ablation study validating our design choices and comparison against leading methods, showcasing the effectiveness of SAViL-Det in diverse and challenging text detection scenarios.

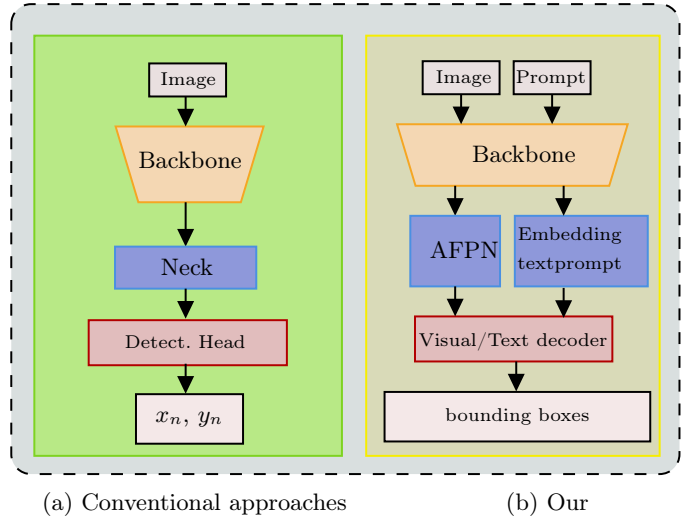
\begin{figure}[htb]
\centering
\begin{flushright}
\centering
\tikzset{every picture/.style={line width=0.75pt}} %set default line width to 0.75pt        

\begin{tikzpicture}[x=0.75pt,y=0.75pt,yscale=-1,xscale=1]
%uncomment if require: \path (0,285); %set diagram left start at 0, and has height of 285

%Shape: Rectangle [id:dp4563191393227042] 
\draw  [fill={rgb, 255:red, 217; green, 223; blue, 223 }  ,fill opacity=1 ][dash pattern={on 4.5pt off 4.5pt}] (1,18) .. controls (1,12.48) and (5.48,8) .. (11,8) -- (328,8) .. controls (333.52,8) and (338,12.48) .. (338,18) -- (338,235.2) .. controls (338,240.73) and (333.52,245.2) .. (328,245.2) -- (11,245.2) .. controls (5.48,245.2) and (1,240.73) .. (1,235.2) -- cycle ;
%Shape: Rectangle [id:dp8482004729309454] 
\draw  [color={rgb, 255:red, 246; green, 237; blue, 0 }  ,draw opacity=1 ][fill={rgb, 255:red, 218; green, 200; blue, 17 }  ,fill opacity=0.19 ] (180,23.91) -- (328,23.91) -- (328,236) -- (180,236) -- cycle ;
%Shape: Rectangle [id:dp8086765273324392] 
\draw  [fill={rgb, 255:red, 233; green, 224; blue, 224 }  ,fill opacity=1 ] (208.48,36.8) -- (247.27,36.8) -- (247.27,53.47) -- (208.48,53.47) -- cycle ;
%Shape: Trapezoid [id:dp7576204351310363] 
\draw  [color={rgb, 255:red, 245; green, 166; blue, 35 }  ,draw opacity=1 ][fill={rgb, 255:red, 244; green, 215; blue, 179 }  ,fill opacity=1 ] (311.9,68.63) -- (302.58,99.7) -- (204.88,99.7) -- (195.56,68.63) -- cycle ;
%Straight Lines [id:da8840069681032643] 
\draw    (227.87,53.47) -- (227.87,65.63) ;
\draw [shift={(227.87,68.63)}, rotate = 270] [fill={rgb, 255:red, 0; green, 0; blue, 0 }  ][line width=0.08]  [draw opacity=0] (8.93,-4.29) -- (0,0) -- (8.93,4.29) -- cycle    ;
%Straight Lines [id:da06347193732391032] 
\draw    (224.87,99.7) -- (224.87,111.85) ;
\draw [shift={(224.87,114.85)}, rotate = 270] [fill={rgb, 255:red, 0; green, 0; blue, 0 }  ][line width=0.08]  [draw opacity=0] (8.93,-4.29) -- (0,0) -- (8.93,4.29) -- cycle    ;
%Shape: Rectangle [id:dp7698642273396594] 
\draw  [color={rgb, 255:red, 74; green, 144; blue, 226 }  ,draw opacity=1 ][fill={rgb, 255:red, 142; green, 152; blue, 218 }  ,fill opacity=1 ] (202.25,114.85) -- (247.5,114.85) -- (247.5,150.47) -- (202.25,150.47) -- cycle ;
%Shape: Rectangle [id:dp5991788265705664] 
\draw  [color={rgb, 255:red, 178; green, 0; blue, 0 }  ,draw opacity=1 ][fill={rgb, 255:red, 225; green, 189; blue, 189 }  ,fill opacity=1 ] (204.61,165.63) -- (304.15,165.63) -- (304.15,188.36) -- (204.61,188.36) -- cycle ;
%Straight Lines [id:da6872775496583403] 
\draw    (224.87,150.47) -- (224.87,162.63) ;
\draw [shift={(224.87,165.63)}, rotate = 270] [fill={rgb, 255:red, 0; green, 0; blue, 0 }  ][line width=0.08]  [draw opacity=0] (8.93,-4.29) -- (0,0) -- (8.93,4.29) -- cycle    ;
%Straight Lines [id:da7395867760191595] 
\draw    (255.67,188.36) -- (255.67,200.52) ;
\draw [shift={(255.67,203.52)}, rotate = 270] [fill={rgb, 255:red, 0; green, 0; blue, 0 }  ][line width=0.08]  [draw opacity=0] (8.93,-4.29) -- (0,0) -- (8.93,4.29) -- cycle    ;
%Shape: Rectangle [id:dp5710615518437019] 
\draw  [fill={rgb, 255:red, 233; green, 224; blue, 224 }  ,fill opacity=1 ] (254.38,36.8) -- (293.16,36.8) -- (293.16,53.47) -- (254.38,53.47) -- cycle ;
%Straight Lines [id:da31883245659184545] 
\draw    (273.77,53.47) -- (273.77,65.63) ;
\draw [shift={(273.77,68.63)}, rotate = 270] [fill={rgb, 255:red, 0; green, 0; blue, 0 }  ][line width=0.08]  [draw opacity=0] (8.93,-4.29) -- (0,0) -- (8.93,4.29) -- cycle    ;
%Shape: Rectangle [id:dp8564365024523826] 
\draw  [color={rgb, 255:red, 74; green, 144; blue, 226 }  ,draw opacity=1 ][fill={rgb, 255:red, 142; green, 152; blue, 218 }  ,fill opacity=1 ] (254.96,114.85) -- (322,114.85) -- (322,150.47) -- (254.96,150.47) -- cycle ;
%Straight Lines [id:da39422964225390755] 
\draw    (279.58,99.7) -- (279.58,111.85) ;
\draw [shift={(279.58,114.85)}, rotate = 270] [fill={rgb, 255:red, 0; green, 0; blue, 0 }  ][line width=0.08]  [draw opacity=0] (8.93,-4.29) -- (0,0) -- (8.93,4.29) -- cycle    ;
%Straight Lines [id:da5629883035781535] 
\draw    (279.58,151.23) -- (279.58,163.39) ;
\draw [shift={(279.58,166.39)}, rotate = 270] [fill={rgb, 255:red, 0; green, 0; blue, 0 }  ][line width=0.08]  [draw opacity=0] (8.93,-4.29) -- (0,0) -- (8.93,4.29) -- cycle    ;
%Shape: Rectangle [id:dp13075898108430128] 
\draw  [color={rgb, 255:red, 126; green, 211; blue, 33 }  ,draw opacity=1 ][fill={rgb, 255:red, 184; green, 233; blue, 134 }  ,fill opacity=1 ] (12.63,23.91) -- (167.76,23.91) -- (167.76,236) -- (12.63,236) -- cycle ;
%Shape: Rectangle [id:dp7454133106118326] 
\draw  [fill={rgb, 255:red, 233; green, 224; blue, 224 }  ,fill opacity=1 ] (70.55,37.93) -- (109.33,37.93) -- (109.33,54.61) -- (70.55,54.61) -- cycle ;
%Shape: Trapezoid [id:dp4614107504991096] 
\draw  [color={rgb, 255:red, 245; green, 166; blue, 35 }  ,draw opacity=1 ][fill={rgb, 255:red, 244; green, 215; blue, 179 }  ,fill opacity=1 ] (128.26,69.76) -- (116.79,108) -- (65.09,108) -- (53.62,69.76) -- cycle ;
%Straight Lines [id:da7214216517224465] 
\draw    (88.94,54.61) -- (88.94,66.76) ;
\draw [shift={(88.94,69.76)}, rotate = 270] [fill={rgb, 255:red, 0; green, 0; blue, 0 }  ][line width=0.08]  [draw opacity=0] (8.93,-4.29) -- (0,0) -- (8.93,4.29) -- cycle    ;
%Shape: Boxed Line [id:dp3162397883232122] 
\draw    (88.94,108) -- (88.94,127.39) ;
\draw [shift={(88.94,130.39)}, rotate = 270] [fill={rgb, 255:red, 0; green, 0; blue, 0 }  ][line width=0.08]  [draw opacity=0] (8.93,-4.29) -- (0,0) -- (8.93,4.29) -- cycle    ;
%Shape: Rectangle [id:dp8411990697543927] 
\draw  [color={rgb, 255:red, 74; green, 144; blue, 226 }  ,draw opacity=1 ][fill={rgb, 255:red, 142; green, 152; blue, 218 }  ,fill opacity=1 ] (53.32,130.39) -- (124.56,130.39) -- (124.56,151.61) -- (53.32,151.61) -- cycle ;
%Shape: Rectangle [id:dp6944349194391921] 
\draw  [color={rgb, 255:red, 178; green, 0; blue, 0 }  ,draw opacity=1 ][fill={rgb, 255:red, 225; green, 189; blue, 189 }  ,fill opacity=1 ] (40.04,167.52) -- (140,167.52) -- (140,190.26) -- (40.04,190.26) -- cycle ;
%Straight Lines [id:da827714420987588] 
\draw    (88.94,151.61) -- (88.94,163.77) ;
\draw [shift={(88.94,166.77)}, rotate = 270] [fill={rgb, 255:red, 0; green, 0; blue, 0 }  ][line width=0.08]  [draw opacity=0] (8.93,-4.29) -- (0,0) -- (8.93,4.29) -- cycle    ;
%Straight Lines [id:da8169617294465557] 
\draw    (88.94,189.5) -- (88.94,201.66) ;
\draw [shift={(88.94,204.66)}, rotate = 270] [fill={rgb, 255:red, 0; green, 0; blue, 0 }  ][line width=0.08]  [draw opacity=0] (8.93,-4.29) -- (0,0) -- (8.93,4.29) -- cycle    ;

% Text Node
\draw (15,253) node [anchor=north west][inner sep=0.75pt]   [align=left] {{\footnotesize (a) Conventional approaches}};
% Text Node
\draw (249,253) node [anchor=north west][inner sep=0.75pt]   [align=right] {{\footnotesize (b) Our}};
% Text Node
\draw (210.52,126.16) node [anchor=north west][inner sep=0.75pt]   [align=left] {{\footnotesize AFPN}};
% Text Node
\draw  [fill={rgb, 255:red, 244; green, 232; blue, 232 }  ,fill opacity=1 ]  (189.26,203.74) -- (323.26,203.74) -- (323.26,228.74) -- (189.26,228.74) -- cycle  ;
\draw (256.26,216.24) node   [align=left] {{\scriptsize bounding boxes}};
% Text Node
\draw (205.0,172.0) node [anchor=north west][inner sep=0.75pt]   [align=left] {{\scriptsize Visual/Text decoder}};
% Text Node
\draw (232.0,78.0) node [anchor=north west][inner sep=0.75pt]   [align=left] {{\footnotesize Backbone}};
% Text Node
\draw (254,119) node [anchor=north west][inner sep=0.75pt]   [align=left] {\begin{minipage}[lt]{40.83pt}\setlength\topsep{0pt}
\begin{center}
{\scriptsize Embedding }\\{\scriptsize textprompt}
\end{center}

\end{minipage}};
% Text Node
\draw (74.94,134.88) node [anchor=north west][inner sep=0.75pt]  [font=\footnotesize] [align=left] {Neck};
% Text Node
\draw  [fill={rgb, 255:red, 244; green, 232; blue, 232 }  ,fill opacity=1 ]  (63.89,204.87) -- (117.89,204.87) -- (117.89,229.87) -- (63.89,229.87) -- cycle  ;
\draw (90.89,217.37) node   [align=left] {{\small {$x_{n}$}, {$y_{n}$}}};
% Text Node
\draw (56.76,173.63) node [anchor=north west][inner sep=0.75pt]  [font=\scriptsize] [align=left] {Detect. Head};

% Text Node
\draw (63.59,81.18) node [anchor=north west][inner sep=0.75pt]  [font=\small] [align=left] {{\footnotesize Backbone}};
% Text Node
\draw (74.76,40.85) node [anchor=north west][inner sep=0.75pt]   [align=left] {{\scriptsize Image}};
% Text Node
\draw (213.7,39.85) node [anchor=north west][inner sep=0.75pt]   [align=left] {{\scriptsize Image}};
% Text Node
\draw (253.94,39.85) node [anchor=north west][inner sep=0.75pt]   [align=left] {{\scriptsize Prompt}};

\end{tikzpicture}
\caption{ Our method versus text detection identification systems.}
\label{fig:fih1}
\end{flushright}
\end{figure}

%%%%%%%%%%%%%%%%%%%%%%%%%%%%%%%%%%%%%%%%
\vspace{-3mm}
\section{\textbf{Related Work}}
\label{sec:relatedWork}
\vspace{-2mm}
\noindent{\textbf{Open-scene text detection.}}
%\vspace{-1mm}
Open-scene text detection can be classified into regression-based and segmentation-based methods. Regression-based methods directly predict text geometry, with advancements towards handling irregular shapes using representations like quadrilaterals or Bezier curves. Segmentation-based approaches  treat text detection as a pixel classification problem, where the goal is to classify each pixel in the image as either belonging to a text instance or to the background. After obtaining the pixel-level classification, a post-processing step is typically applied to group the text pixels into individual text instances and generate the final bounding boxes or contours. For instance, Maoyuan \textit{et al.} \cite{deepsolo++} proposed a unified framework that combines text detection and recognition by using a single Transformer decoder. This approach introduces explicit point queries and a Bezier curve representation, offering faster convergence and reduced annotation costs. Xiang \textit{et al.} \cite{PSENet} introduced a segmentation-based text detection method that employs a progressive scale expansion algorithm, enabling the detection of arbitrarily shaped text in natural scenes. DBNet~\cite{DBNet} integrated Differentiable Binarization (DB) and Adaptive Scale Fusion (ASF) to improve segmentation performance. By embedding binarization within the network, DB enhances text boundary detection, while ASF dynamically fuses multi-scale features, improving robustness to scale variations and complex text shapes. PixelLink~\cite{PixelLink} introduced an approach that links pixels within the same text instance to form Connected Components (CCs). It utilizes a DNN with VGG16 as the feature extractor, predicting both text/non-text labels and pixel links. By connecting positive pixels through predicted links, it generates  connected components that represent text instances, from which bounding boxes are extracted.\\

\noindent{\textbf{Cross-modal text detection.}}
%Transformer-based scene text detection methods, while popular for predicting text boundaries as polygon or Bezier curve points, often suffer from limitations in training efficiency and detection robustness due to their reliance on coarse positional query modeling and point label forms that inherently encode reading order. To overcome these challenges, DPText-DETR is proposed, a novel Dynamic Point Text DEtection TRansformer network that directly utilizes explicit point coordinates to generate positional queries which are then dynamically updated throughout the network. Furthermore, the introduction of an Enhanced Factorized Self-Attention module with circular shape guidance improves the spatial inductive bias, and a practical positional label form enhances detection robustness, culminating in state-of-the-art performance on standard benchmarks and a newly introduced Inverse-Text dataset for evaluating robustness to text orientation
Cross-modal text detection represents a new paradigm shift in the field by integrating information from multiple modalities beyond the raw visual input. These modalities commonly include the surrounding context, and even the textual content already present within the image. The core principle behind cross-modal text detection is that by synergistically combining these diverse sources of information, a more robust and accurate detection of text can be achieved, especially in challenging conditions where visual cues alone might be ambiguous or insufficient. To overcome these challenges, the work in~\cite{TCM} leverages the pre-trained CLIP model for scene text detection without requiring additional pretraining. By utilizing visual and language prompt generators with cross-modal interaction, \cite{TCM} efficiently aligns image and text embeddings, enhancing text detection capabilities. The method shows strong domain adaptation and few-shot learning performance, achieving significant improvements with minimal labeled data. Xue {\it et al.} ~\cite{Xue} introduced a weakly supervised pre-training technique for scene text detection by jointly learning visual and textual features. It uses a character-aware text encoder to capture relevant text instance information, and a visual-textual decoder to enhance visual representations by modeling interactions between image and text. VLPT-STD framework~\cite{VLPT-STD} integrates visual and textual representations through a multi-component architecture. A text encoder processes sub-word tokens using a cross-modal encoder that facilitates the interactions between the visual and text embeddings through cross-attention. 

In contrast to the aforementioned and existing methods (Fig. \ref{fig:fih1}), we propose a novel cross-modal pre-training paradigm drawing inspiration from recent vision-language pre-training techniques. Our framework features a language-vision decoder that adaptively propagates fine-grained semantic information from textual to visual features, moving beyond traditional encoder-decoder attention mechanisms.

%%%%%%%%%%%%%%%%%%%%%%%%%%%%%%%%%%%%%%%%%%
\vspace{-3mm}
\section{The Proposed Approach: SAViL-Det}
\label{sec:proposed}
\begin{figure*}[ht]
\resizebox{\textwidth}{!}{\input{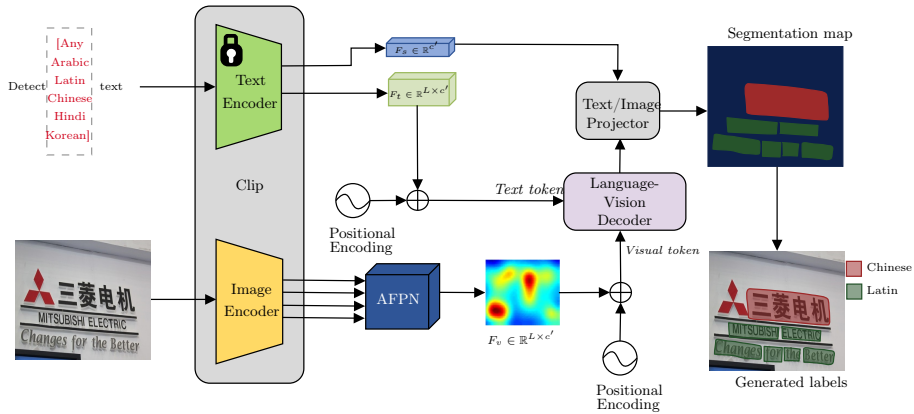}}
\caption{SAViL-Det architecture. Given an input image and a text prompt, our approach begins by leveraging CLIP with a ResNet-50 backbone to extract representative features from both the image and the input text prompt. Asymptotic Feature Pyramid Network (AFPN) is then used for multi-scale visual feature fusion.}
\label{fig:overall}
\end{figure*}

Given an input image and a text prompt, we aim to detect the text present in the image and determine its script. To achieve this, we propose a novel approach SAViL-Det, as illustrated in Fig. \ref{fig:overall}. The different components of our approach are described  below.

%Our method begins by leveraging CLIP with a ResNet-50 backbone to extract representative features from both the image and the input text prompt.

\vspace{-1cm}
\subsection{\textbf{Visual and Text Feature Extraction}} 
%As shown in Fig. \ref{fig:overall}, the input to our text detector consists of an image \( I \) and a text prompt \( T \). In this section, we describe the process of extracting visual and textual features from these inputs, using ResNet-50 for the image and the CLIP text encoder for the text.

\subsubsection{\textbf{Image encoder:}}

Given an input image \( I \in \mathbb{R}^{H \times W \times 3} \), we extract visual features from all stages of ResNet-50, excluding the first stage. Specifically, features are obtained from the 2nd, 3rd, 4th, and 5th stages, represented as \( F_{v2} \in \mathbb{R}^{\frac{H}{4} \times \frac{W}{4} \times C_2} \), \( F_{v3} \in \mathbb{R}^{\frac{H}{8} \times \frac{W}{8} \times C_3} \), \( F_{v4} \in \mathbb{R}^{\frac{H}{16} \times \frac{W}{16} \times C_4} \), and \( F_{v5} \in \mathbb{R}^{\frac{H}{32} \times \frac{W}{32} \times C_5} \), respectively. Here, \( C_i \) denotes the feature dimension, while \( H \) and \( W \) represent the height and width of the original image. The ResNet-50 backbone remains trainable during the training stage.

\vspace{-1mm}
\subsubsection{\textbf{Text encoder:}}
Given an input text prompt \( T\in \mathbb{R}^L \), we utilize the CLIP text encoder to extract the text features \( F_t \in \mathbb{R}^{L \times C} \). The encoder processes the text using a lowercased Byte Pair Encoding (BPE) representation, with a vocabulary size of 49,152. The sequence is bracketed with [SOS] and [EOS] tokens to mark the beginning and the end of the text. The activations from the highest layer of the encoder on the [EOS] token are then transformed into the global textual representation \( F_s \in \mathbb{R}^{C'} \). Here, \( C \) and \( C' \) represent the feature dimensions, and \( L \) denotes the length of the input prompt. Since the CLIP text encoder is sensitive to language phrasing, we used three slightly different input text prompts (see Tab \ref{tab:prompts}). Each prompt guides the model's attention differently—some emphasise raw visual content ("image") while others emphasize the broader semantic context ("scene"). This diversity helps to capture a wider range of text instances and improves detection reliability. During training, the CLIP text encoder parameters remain frozen.

\vspace{-3mm}
\begin{table}[]
\centering
\caption{Predefined text prompts to detect full text scripts within input images.}
\begin{tabular}{cl}
\hline
Prompt No. & Prompts                             \\ \hline
P1         & Detect Any text in the image.       \\  
P2         & Where is text located in the scene? \\
P3         & Detect Any text in the scene.       \\ \hline
\end{tabular}
\label{tab:prompts}
\end{table}
%%%%%%%%%%%%%%%%%%%%%%%%%%%%%%%%%%
\vspace{-6mm}
\subsection{\textbf{Asymptotic Feature Pyramid Network}}

To fuse the multi-scale visual features, we integrate the Asymptotic Feature Pyramid Network (AFPN) ~\cite{yang2023afpn}, which enhances upon FPN by directly fusing non-adjacent features and reducing visual feature loss. Unlike FPN, which fuses only adjacent feature levels, AFPN progressively combines features from different levels, starting from low-level features and gradually incorporating high-level features. The resulted multiple visual features of the global textual representation \( F_0, F_1, F_2, F_3 \) represent the multi-scale global textual representation. The fusion process begins by fusing adjacent low-level features \( F_0 \) and \( F_1 \) as follows:
\begin{equation}
F_{v1} = \textit{Conv}([F_0, \textit{Up}(F_1)])
\end{equation}
where \textit{Up(.)} denotes 1× upsampling. Then, the fused feature \( F_{v1} \) is upscaled and combined with the mid-level feature \( F_2\):
\begin{equation}
F_{v2} = \textit{Conv}([F_{v1}, \textit{Up}(F_2)])
\end{equation}
Finally, the highest-level feature \( F_3 \) is incorporated into the fusion process, resulting in the final fused feature map \( F_{v3} \):
\begin{equation}
F_{v3} = \text{Conv}(\textit{Up}(F_{v2}), F_3)
\end{equation}
To ensure that useful features from all levels are effectively combined, an adaptive spatial fusion operation is applied at each fusion step. This operation assigns spatial weights \( \alpha, \beta, \gamma, \delta \) to each feature, allowing the network to prioritize the most relevant features from different levels:

\begin{equation}
F_{v} = \alpha \cdot F_{v0} + \beta \cdot F_{v1} + \gamma \cdot F_{v3}
\end{equation}

\noindent where \( \alpha, \beta, \gamma, \delta \) are learnable weights that sum to 1. This weighted fusion ensures a better preservation of both fine details of low-level features and rich semantics from high-level features.

%%%%%%%%%%%%%%%%%%%%%%%%%%%%%%%%%%%%%%%%%%
\subsection{\textbf{Cross-Modal Decoder Transformer}}
We noticed that existing methods leveraging transformers to build networks with encoder-decoder attention mechanisms tend to neglect the integration of textual prompts. Instead, these methods rely heavily on visual context information and depend solely on self-attention. We propose a novel language-vision decoder that adaptively propagates fine-grained semantic information from textual features to visual features, as depicted in Fig \ref{fig:overall}. We elaborate on the language-vision decoder in the following sections.

Let \( F_t \in \mathbb{R}^{(h \times w) \times d} \) and \( F_v \in \mathbb{R}^{N \times C} \) represent the input tensors for textual and visual features, respectively, which can provide sufficient textual information corresponding to visual features. To capture positional information, fixed sinusoidal spatial positional encodings are added to \( F_v \) and \( F_t \).

\begin{figure*}[ht]
\centering
\input{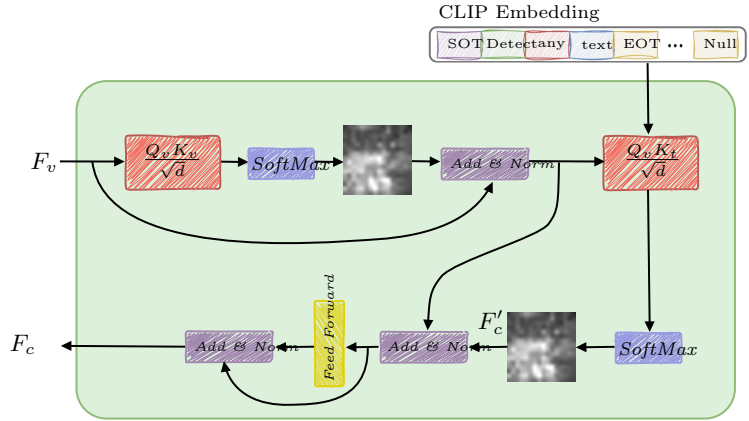}
\caption{The decoder takes visual features $F_t \in \mathbb{R}^{N \times C}$ and textual features $F_t$, applying multi-head self-attention (MHSA) to update the visual features, then multi-head cross-attention (MHCA) to integrate semantic information. The final multi-modal features $F_c$ are produced via a multi-layer perceptron (MLP) with residual connections.}
\label{fig:decoder}
\end{figure*}

The vision-language decoder, composed of $n$ layers, generates multi-modal features $F_c \in R^{N \times C}$ using a transformer architecture.  Each layer includes a multi-head self-attention (MHSA) layer, a multi-head cross-attention (MHCA) layer, and a feed-forward network (see Fig  \ref{fig:decoder}).
In the MHSA layer, the visual feature $F_v$ is processed to capture global context, evolving into $F'_v$. The self-attention mechanism uses queries, keys, and values derived from $F_v$, with the attention operation computed as:
\begin{equation}
\text{MHSA}(Q_v, K_v, V_v) = \text{softmax}\left( \frac{Q_v K_v^T}{\sqrt{d_k}} \right) V_v
\end{equation}
Next, the MHCA layer incorporates semantic information from textual features $F_t$, refining the visual features $F'_v$ into $F'_c$. Finally, a multi-layer perceptron (MLP) with residual connections produces the final multi-modal features $F_c$.

%%%%%%%%%%%%%%%%%%%%%%%%%%%%%%%%%%%%%%%%%%
\subsection{\textbf{Text/Image Projector}}
We incorporated a text/image projector layer into our framework, which precisely aligns the textual features with the corresponding pixel-level visual features. The image and text projectors are used to transform the visual and textual features \(F_c\) and \(F_s\), respectively, as follows:
\begin{equation}
z_v = Upsample(F_c) W_v + b_v, \quad z_t = F_s W_t + b_t
\end{equation}

\noindent where \( \mathbf{z}_t \in \mathbb{R}^D \) and \( \mathbf{z}_v \in \mathbb{R}^{N \times D} \), with \( N = \frac{H}{4} \times \frac{W}{4} \). Here, \( \text{Upsample} \) refers to a 4× upsampling operation, and \(W_v \) and \( W_t \) are learnable weight matrices that project \(F_c \) and \(F_s \) into a shared feature space of dimension \( D \), while \( b_v \) and \(b_t \) are learnable biases.
Once the textual feature \( \mathbf{z}_t \) and pixel-level visual features \( \mathbf{z}_v \) are transformed, a contrastive loss function is employed to optimize the alignment between the two modalities. The objective is to maximize the similarity between corresponding \( \mathbf{z}_t \) and \( \mathbf{z}_v \) while minimizing the similarity with non-corresponding features. The similarity between the text and visual features is measured via their dot product, and the contrastive loss is formulated as:
\begin{equation}
\mathcal{L}_{\text{contrastive}} = - \sum_{i \in P} \log \sigma(\mathbf{z}_t \cdot \mathbf{z}_v) - \sum_{i \in N} \log \left( 1 - \sigma(\mathbf{z}_t \cdot \mathbf{z}_v) \right)
\end{equation}

\noindent where \( P \) and \( N \) correspond to the positive and negative pairs in the ground truth, and \( \sigma(\cdot) \) denotes the sigmoid function.

Finally, to obtain the aligned output, we reshape the similarity scores \( \sigma(\mathbf{z}_t \cdot \mathbf{z}_v) \) to the resolution \( \frac{H}{4} \times \frac{W}{4} \) and subsequently upsample the result to match the original image size, thus generating the final aligned feature map for downstream tasks.

%%%%%%%%%%%%%%%%%%%%%%%%%%%%%%%%%%%%%%%%%%
\vspace{-2mm}
\section{\textbf{Experimental Analysis}}
\label{sec:experiments}

\vspace{-2mm}

%%%%%%%%%%%%%%%%%%%%%%%%%%%%%%%%%%%%%%%%%%
\subsection{\textbf{Datasets}}
The proposed approach is evaluated on two datasets, each exhibiting diverse complexities and challenges, including texts with varying orientations, sizes, and languages. %The following subsections offer concise descriptions of the utilized datasets.

\noindent \textbf{Multi-lingual Text 2019} (MLT-2019) dataset~\cite{MLT2019} is a large-scale multilingual collection designed for scene text detection, featuring 20,000 images. It covers 10 languages across 7 scripts, with 10,000 images allocated for training and 10,000 for testing. Languages include \textit{Arabic, Bangla, Chinese, Devanagari, English, French, German, Italian, Japanese}, and \textit{Korean}. Each language is represented equally in both the training and testing sets.

\noindent \textbf{CTW1500 }dataset~\cite{CTW1500} is designed for OCR tasks with a focus on curved text. It comprises 1,000 training images and 500 testing images. Text instances are annotated at the text line level using polygons with 14 vertices. This detailed annotation helps in the accurate recognition of curved text in various contexts.
%%%%%%%%%%%%%%%%%%%%%%%%%%%%%%%%%%%%%%%%%%
\vspace{-2mm}
\subsection{\textbf{Evaluation Metrics}}
\vspace{-2mm}
The intersection over union (IoU) metric is used to assess the accuracy of the model in detecting text regions and language identification. Precision (P), recall (R), and F-measure (F) are calculated for comparative analysis, as commonly reported in other works, e.g.~\cite{karatzas2013icdar}. To ensure fair comparisons, text BBoxes or text labelled as "\#\#\#" are ignored in both training and testing across all datasets.

\vspace{-3mm}
\subsection{\textbf{Implementation Details}}
\vspace{-2mm}
In this study, we used ResNet-50 as the backbone for the CLIP image encoder, with input images resized to 512 × 512 pixels. The input text prompts, which include start and end tokens, are limited to a maximum of 77 tokens. The language-vision decoder consists of 8 attention heads per layer, and the feedforward hidden dimension is set to 1024, with xFormers \cite{xFormers2022} utilising low-rank approximations to reduce GPU memory consumption. For training, we set the batch size to 32 to strike a balance between computational efficiency and convergence speed. The model was trained on four NVIDIA Tesla V100 GPUs using distributed training. We employed the Adam optimizer with a learning rate of 0.0001 and a weight decay of 1e-5, with a learning rate decaying by a factor of 0.1 every 10 epochs. The training lasted for 110 epochs. Data augmentation techniques, including random cropping and flipping, were applied. All experiments were implemented in PyTorch with mixed precision training to optimize memory usage.

%%%%%%%%%%%%%%%%%%%%%%%%%%%%%%%%%%%%%%%%%%%
\vspace{-2mm}
\subsection{\textbf{Results and Comparison with State-of-the-Art (SOTA)}}

Tables \ref{tab:MLT2019_res} and \ref{tab:CTW1500_res} show the performance of SAViL-Det and a comparison with other text detection methods on MLT 2019 and CTW1500 datasets. Our method achieves the highest F-score on both datasets, proving its effectiveness in detecting complex text instances. Specifically, SAViL-Det achieved an F-score of 84.8\% on MLT 2019. On CTW1500, our model also outperforms other approaches, reaching 90.2\% in the F-score, slightly surpassing MixNet. These results demonstrate the robustness of SAViL-Det in handling multi-oriented and long text instances across different datasets.

\begin{figure*}[htb]
\centering
\includegraphics[width=1.0\textwidth]{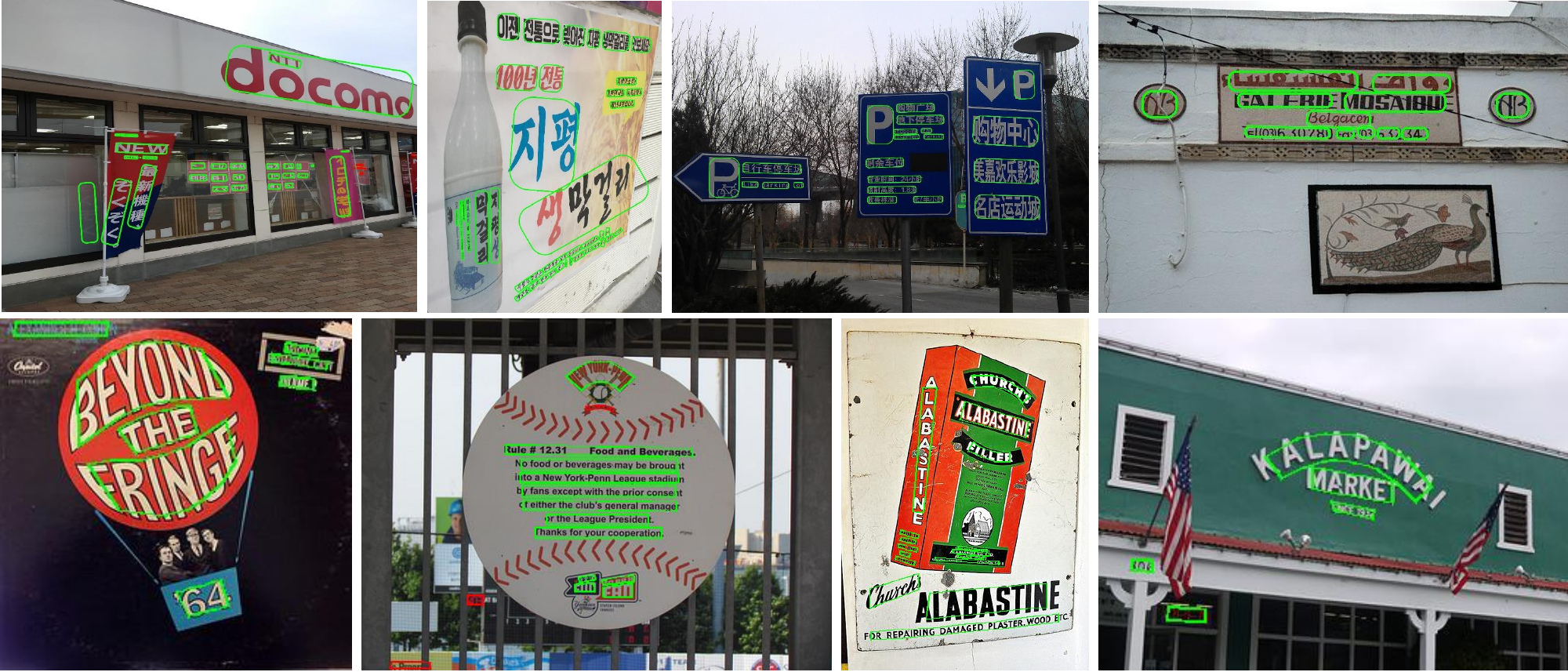}
\caption{Visualization of text detection results on MLT2019 dataset (Top) and CTW1500 dataset (bottom). The red bboxes show the very few false detections.}
\label{fig:results}
\end{figure*}

The improvements achieved by SAViL-Det can be attributed to its novel architecture incorporating a vision-language decoder and a text-to-pixel contrastive learning mechanism. Our text detection method allows for a more effective alignment between visual text features and their corresponding textual descriptions. The vision-language decoder facilitates the propagation of fine-grained semantic information from the text to the pixel level enabling the model to better understand and locate text instances. Furthermore, the text-to-pixel contrastive learning mechanism explicitly encourages the model to learn representations where the textual features are similar to the visual features of the corresponding text pixels and dissimilar to those of the background, leading to more accurate and robust text detection, especially in challenging scenarios with complex layouts and orientations. Fig. \ref{fig:results} shows qualitative results, where our method produces accurate text boundaries with few false positives in difficult cases.

\vspace{-1cm}
\begin{table}[htb]
\centering
\caption{Detection results on MLT2019 dataset.}
\begin{tabular}{lccc}
\hline
\textbf{Method}     & \textbf{Recall} &   \textbf{Precision} & \textbf{F-score} \\ \hline
LOMO ~\cite{lomo}           & 79.8  & 87.8  & 83.6 \\
DPText-DETR-Res50 \cite{DPText-DETR} & 73.7 &  83.0  &  78.1 \\
EAST + PVANET2x ~\cite{east}& 75.1  & 81.6  & 78.2 \\
SRFormer \cite{SRFormer}    & 73.5  &  86.1 & 79.3 \\
Pyramid Context Network     & 81.5  & 84.6  & 83.0 \\
DeepSolo++ \cite{deepsolo++} & 66.9 & 85.1 & 74.9 \\
PixelLink + VGG16 4s  \cite{PixelLink} & 82.3  & 82.1  & 82.1 \\
PixelLink + VGG16 2s \cite{PixelLink} & 82.0  & 85.5  & 83.7 \\ \hline
\textbf{SAViL-Det (Our)} & \textbf{82.6} & \textbf{87.2} & \textbf{84.8}\\ \hline
\end{tabular}
\label{tab:MLT2019_res}
\end{table}

\vspace{-1cm}
\begin{table}[htb]
\centering
\caption{Detection results on CTW1500 dataset.}
\begin{tabular}{lccc}
\hline
\textbf{Method} & \textbf{Recall} & \textbf{Precision} & \textbf{F-score} \\
\hline
TextSnake\cite{Textsnake}& 85.3  &  67.9  &  75.6 \\
PSENet~\cite{PSENet}     & 79.7  &  84.8  &  82.2 \\
DBNet-ResNet50\cite{DBNet}& 83.2  &   88.9  &  86.0 \\
CRAFT \cite{CRAFT}       & 81.1  &  86.0  &  83.5 \\
FAST-B \cite{FAST_2021}  & 80.9  &  87.8  &  84.2 \\
PAN \cite{pan}           & 83.2  &  86.8  &  85.0 \\
FCENet ~\cite{FCENet}    & 82.8  &  87.5  &  85.1 \\
TESTR-Polygon\cite{TESTR}& 82.6  &  92.0  &  87.1 \\
DPText-DETR \cite{DPText-DETR}& 86.2  &  91.7  &  88.8 \\
DeepSolo++ \cite{deepsolo++} & 86.3 & \textbf{92.5} & 89.3 \\
SRFormer \cite{SRFormer} & 89.8  &  89.4  &  89.6 \\
MixNet \cite{mixnet}     & 88.3  &  91.4  &  89.8 \\
\hline
\textbf{SAViL-Det (Our)} & \textbf{89.5} & 90.9 & \textbf{90.2}\\ \hline
\end{tabular}
\label{tab:CTW1500_res}
\end{table}

%%%%%%%%%%%%%%%%%%%%%%%%%%%%%%%%%%%%%%%%%%%%%

%%%%%%%%%%%%%%%%%%%%%%%%%%%%%%%%%%%%%%%%%%%
%\vspace{-1cm}
\subsection{\textbf{Ablation Study}}
%\vspace{-1cm}
%To demonstrate the efficacy of the proposed components, we performed two ablation studies on both the MLT2019 and CTW1500 datasets.

\subsubsection*{\textbf{Vision-language decoder's depth.}}
To investigate the impact of the vision-language decoder's depth on the performance of our SAViL-Det method, we conducted an ablation study by varying the number of transformer layers within the decoder. We evaluated the recall at different Intersection-over-Union thresholds (0.50, 0.60, 0.70, 0.80, and 0.90) on the MLT2019 and CTW1500 datasets. The results are presented in Table \ref{tab:decoder_depth}. We observed that increasing transformer layers from 2 to 3 improved F-score across all IoU thresholds on both the MLT2019 and CTW1500 datasets, with notable gains at F1@50 and F1@90. This enhancement suggests that the additional layer helps refining  multi-modal features and improving the text detection accuracy. However, adding more layers (4 or 5) led to performance declines, likely due to overfitting or diminishing returns in feature refinement. The findings indicate that 3 transformer layers offer an optimal balance for the vision-language decoder in our SAViL-Det method.

\vspace{-3mm}
\begin{table}[h!]
\centering
\caption{Ablation study on vision-language decoder's depth using F-scores at different IoU on MLT2019 and CTW1500 datasets.}
\begin{tabular}{c|c|ccccc}
\hline
Dataset                  & Layers & F1@50& F1@60& F1@70& F1@80 & F1@90   \\ \hline
\multirow{4}{*}{MLT2019} & 2           & 81.9 & 74.5 & 54.3 & 34.8 &  21.2   \\
                         & 3           & 84.8 & 77.3 & 60.9 & 39.6 &  22.4   \\
                         & 4           & 83.2 & 75.9 & 58.3 & 38.1 &  22.9   \\
                         & 5           & 79.9 & 71.1 & 53.3 & 29.8 &  19.7   \\ \hline
                         
\multirow{4}{*}{CTW1500} & 2           & 86.9 & 75.2 & 68.5 & 53.3 &  29.5   \\
                         & 3           & 90.2 & 78.9 & 71.1 & 59.6 &  32.4   \\
                         & 4           & 89.5 & 77.8 & 69.2 & 58.1 &  30.3   \\
                         & 5           & 86.9 & 70.1 & 59.2 & 48.4 &  28.1   \\ \hline
\end{tabular}

\label{tab:decoder_depth}
\end{table}

\vspace{-1cm}
\subsubsection*{\textbf{Impact of text prompt features.}}
In order to investigate the important role of textual features in cross-modal decoder in our framework, we performed another ablation study, in which we completely removed the text input from the decoder. This specific configuration, which we refer to as “No Textual Features”, forces the model to depend only on visual information that is extracted from the CLIP image encoder. The obtained results, compared to the full model (“Yes” textual features), are shown in Table \ref{tab:no_language_ablation}.

\begin{table}[h!]
\centering
\caption{Impact of language input. "Yes" represents the full model with textual features. "No" represents the model without language input to the vision-language decoder.}
\begin{tabular}{cccclccc}
\hline
\multirow{2}{*}{Textual features} & \multicolumn{3}{c}{MLT2019} &  & \multicolumn{3}{c}{CTW1500} \\ \cline{2-4} \cline{6-8} 
             & R       & P      & F-score   &  & R      & P      & F-score   \\ \hline
No           & 80.2    & 82.7   & 81.4      &  & 85.3   & 67.9   & 75.6      \\
Yes          & 82.6    & 87.2   & 84.8      &  & 89.5   & 90.9   & 90.2      \\ \hline
\end{tabular}
\label{tab:no_language_ablation}
\end{table}

Not using the textual features led to a performance descrease. On MLT 2019, the F-score decreased by 3.4\% (84.8\% to 81.4\%). On CTW1500, the decrease was even more significant —14\% (89.6\% to 75.6\%). The larger decline on CTW1500, particularly in precision (89.4\% to 67.9\%), suggests that language input is crucial for handling complex text shapes and reducing false positives. Without textual features, the vision encoder alone struggles with CTW1500’s curved and irregular text instances.

The results confirm that the language component in our cross-modal decoder is essential, providing contextual guidance that enhances both recall and precision, particularly in challenging scenarios. By incorporating textual features, the model better associates visual cues with linguistic concepts, improving localization accuracy.

%%%%%%%%%%%%%%%%%%%%%%%%%%%%%%%%%%%%%%%%%%
\vspace{-4mm}
\section{Conclusion}
\label{sec:conclusion}
\vspace{-2mm}
This paper described SAViL-Det, a semantic-aware vision-language method to address challenges in multi-script and complex scene text detection. Our approach uses CLIP and a novel language-vision decoder with cross-modal attention and contrastive learning to effectively integrate textual prompts with visual features. Experiments confirmed the benefits of this semantic guidance, demonstrating state-of-the-art performance on the MLT 2019 and CTW1500 datasets. This work highlights the value of vision-language integration for text detection. Future directions include exploring text spotting applications and optimizing efficiency.

\vspace{-2mm}
%%%%%%%%%%%%%%%%%%%%%%%%%%%%%%%%%%%%%%%%%%%%%%%%%%%%%%%%%%%%%%%%%%

\bibliographystyle{splncs04}
\bibliography{mybibliography}

\end{document}